\documentclass[times,authoryear]{elsarticle}

\usepackage[utf8]{inputenc}
\usepackage{graphicx}
\usepackage{url}
\graphicspath{ {images/} }
\usepackage{caption}
\usepackage{amsmath}
\usepackage{verbatim}

\usepackage{dingbat} 

\usepackage{subfigure}

\usepackage{orcidlink}

\usepackage[authoryear]{natbib}

\RequirePackage{fix-cm}

\usepackage{tablefootnote}
\usepackage{pdflscape}

\def\tsc#1{\csdef{#1}{\textsc{\lowercase{#1}}\xspace}}
\tsc{WGM}
\tsc{QE}
\tsc{EP}
\tsc{PMS}
\tsc{BEC}
\tsc{DE}

\usepackage{hyperref}
\hypersetup{
    colorlinks=true,
    linkcolor=blue,
    filecolor=magenta,      
    urlcolor=cyan,
    pdftitle={Overleaf Example},
    pdfpagemode=FullScreen,
    }    
\usepackage[normalem]{ulem}

\begin{document}

\title{Multi-Modal Landslide Detection from Sentinel-1 SAR and Sentinel-2 Optical Imagery Using Multi-Encoder Vision Transformers and Ensemble Learning}

\author[1]{Ioannis Nasios}

\ead{ioannis.nasios@nodalpoint.com}

\affiliation[1]{organization={Nodalpoint Systems},
    addressline={Pireos 205}, 
    city={Athens},
    postcode={118 53}, 
    country={Greece}}
    
\begin{frontmatter}

\begin{abstract}
Landslides represent a major geohazard with severe impacts on human life, infrastructure, and ecosystems, underscoring the need for accurate and timely detection approaches to support disaster risk reduction. This study proposes a modular, multi-model framework that fuses Sentinel-2 optical imagery with Sentinel-1 Synthetic Aperture Radar (SAR) data, for robust landslide detection. The methodology leverages multi-encoder vision transformers, where each data modality is processed through separate lightweight pretrained encoders, achieving strong performance in landslide detection. In addition, the integration of multiple models, particularly the combination of neural networks and gradient boosting models (LightGBM and XGBoost), demonstrates the power of ensemble learning to further enhance accuracy and robustness. Derived spectral indices, such as NDVI, are integrated alongside original bands to enhance sensitivity to vegetation and surface changes. The proposed methodology achieves a state-of-the-art F1 score of 0.919 on landslide detection, addressing a patch-based classification task rather than pixel-level segmentation and operating without pre-event Sentinel-2 data, highlighting its effectiveness in a non-classical change detection setting. It also demonstrated top performance in a machine learning competition, achieving a strong balance between precision and recall and highlighting the advantages of explicitly leveraging the complementary strengths of optical and radar data. The conducted experiments and research also emphasize scalability and operational applicability, enabling flexible configurations with optical-only, SAR-only, or combined inputs, and offering a transferable framework for broader natural hazard monitoring and environmental change applications. Full training and inference code can be found in \url{https://github.com/IoannisNasios/sentinel-landslide-cls}.
\end{abstract}

\end{frontmatter}

\section{Introduction}
\label{sec:introduction}
The accurate detection of landslides is a critical task in natural hazard monitoring, as landslides represent a major threat to human lives, infrastructure, and ecosystems. They are frequently triggered by external events such as earthquakes, prolonged rainfalls, or extreme weather conditions, and their rapid onset often limits opportunities for timely intervention. Reliable methods for landslide mapping are therefore essential to support risk mitigation, emergency response, and long-term resilience planning.

Landslides represent a major natural hazard with significant social and economic impacts worldwide. Globally, they cause substantial damage to infrastructure and economic systems, with annual losses estimated at around 20 billion USD, largely due to impacts on transportation networks and road infrastructure \citep{baby2026economic}. Beyond direct repair costs, landslides also generate indirect economic effects such as traffic disruption and supply chain interruptions, further affecting regional development. At a broader scale, approximately 13\% of the global land area is currently classified as highly susceptible to landslides, particularly in major mountainous regions such as the Andes, Alps, and Himalayas \cite{duan2025global}. Climate change is expected to further increase landslide susceptibility in many regions, highlighting the growing importance of improved monitoring and risk assessment approaches.

Satellite remote sensing (RS) offers a scalable means to detect and map landslides rapidly over broad areas, with optical sensors providing high interpretability of surface reflectance changes and Synthetic Aperture Radar (SAR) enabling observations regardless of daylight or cloud cover. In their comprehensive review, \cite{novellino2024mapping} analyzed 291 studies worldwide and reported a 600\% surge in publications after 2014, a trend largely driven by the availability of Sentinel-1 and Sentinel-2 data and the rapid rise of artificial intelligence (AI), which has surpassed all other techniques since 2020. In this context, Sentinel-2's MultiSpectral Instrument (MSI, Level-2A) supplies various optical bands with 10, 20 or 60 m resolution out of which the 3 RGB and the NIR band (all with 10 m resolution) are useful for detecting vegetation loss, bare‐soil exposure, and moisture-related spectral changes, while Sentinel-1 C-band SAR provides VV/VH backscatter and coherence information sensitive to surface roughness and structural disturbance. Prior studies have demonstrated the value of each modality independently for landslide mapping \citep{kyriou2018assessing,lacroix2018use,notti2022semi,santangelo2022exploring,monopoli2024landslide}.

Optical imagery from Sentinel-2 has been extensively utilized for post-failure landslide detection, change analysis, and scar identification owing to its high spatial resolution and straightforward interpretability \citep{qu2021post}. Typical approaches rely on pre- and post-event change detection in spectral reflectance or vegetation indices, such as the  Normalized Difference Vegetation Index (NDVI), to delineate fresh scarps and debris zones. Recent advances have demonstrated accurate and rapid event mapping from freely available Sentinel-2 data, both through multi-temporal optical analyses \citep{satriano2023landslides} and robust change detection techniques \citep{coluzzi2025rapid}.

SAR imagery, including Sentinel-1 acquisitions, provides a valuable alternative by penetrating clouds and acquiring data regardless of illumination. SAR-based techniques, such as interferometric coherence analysis and polarimetry, have been effectively applied to detect landslides, assess post-failure stability, and monitor subtle ground deformations \citep{ohki2020landslide, di2016landslide}. Multi-temporal SAR imagery enables change detection that complements optical analyses, providing a robust framework for both small-scale and large-scale landslide monitoring. Sentinel-1 SAR has proven effective under cloud or nighttime conditions, using amplitude, coherence loss, and multitemporal statistics to capture ground disturbance. Multiple investigations report robust relationships between co-event coherence/backscatter changes and mapped landslides \citep{tzouvaras2020small,burrows2020systematic}.

The evolution of RS technologies has facilitated more robust landslide detection and longitudinal monitoring, maintaining high performance under heterogeneous environmental conditions and diverse observational parameters. \cite{liang2025landslide} demonstrated the potential of single-temporal post-event polarimetric SAR imagery for accurate landslide mapping through a deep learning method that exploits morphological characteristics of landslide features. Their approach effectively captured landslide shapes and boundaries using only post-event data, achieving high accuracy even with dual-polarized Sentinel-1 imagery, underscoring the utility of SAR data for rapid response and cloud-independent mapping. Complementarily, \cite{li2025mapping} addressed the challenge of detecting small, forest-covered landslides by introducing a refined interferometric phase processing framework that enhances localized phase detail. Together, these studies highlight the growing role of advanced SAR-based deep learning and interferometric methods in extending landslide detection capabilities beyond traditional optical limitations, especially in cloud-prone or densely vegetated regions.

Advances in RS and machine learning (ML) have substantially improved the detection, mapping, and monitoring of landslides, transitioning from localized assessments to broad regional and global scales. 
Integrated approaches that combine satellite observations with in situ information have demonstrated the value of multi-source data for updating inventories and characterizing rainfall-induced failures, as shown by \cite{miele2022sar}, who merged SAR data with field surveys to enhance shallow landslide mapping. Similarly, ML methods are increasingly employed to assess landslide hazards and identify risk hotspots, with \cite{sundriyal2024integrated} illustrating how RS–driven ML frameworks can support detailed hazard evaluations in complex mountainous environments. At larger scales, heterogeneous ensemble deep-learning models have proven effective for automating global landslide detection, offering robust classification across diverse geomorphic and imaging conditions \citep{ganerod2024automating}. More recently, the development of lightweight multimodal architectures, such as dual-stream attention networks, has enabled real-time landslide monitoring from both optical and SAR imagery, underscoring the growing relevance of efficient fusion-based approaches for operational applications \citep{dhayal2025lightweight}. Building upon this trajectory of multimodal and ML-enabled innovation, the present study advances landslide detection by integrating Sentinel-1 SAR and Sentinel-2 optical data within a multi-encoder transformer and ensemble learning framework.

Recent studies have explored strategies to reduce the reliance of deep learning models on large labeled datasets for landslide detection. Unsupervised approaches based on convolutional autoencoders can extract high-level features from Sentinel-2 imagery and topographic data, enabling effective landslide detection without labeled samples \citep{shahabi2021unsupervised}. Similarly, contrastive self-supervised learning methods can learn transferable representations from large volumes of unlabeled satellite imagery, achieving competitive performance while using only a small fraction of labeled data \citep{ghorbanzadeh2024contrastive}. These advances highlight the growing potential of representation learning for leveraging unlabeled Earth observation (EO) data in geohazard monitoring.

Fusing Sentinel-1 and Sentinel-2 can improve robustness and reduce errors of oversight, failures, or neglects relative to single-sensor methods, particularly in heterogeneous terrain or cloudy conditions. Recent works have explored optical-SAR integration for rapid event mapping and object-based or learning-based classification, underscoring the practical benefits of multi-source data in operational workflows \citep{ghorbanzadeh2020application,kyriou2020landslide,prodromou2023rapid}. At the same time, community efforts to build larger, more diverse benchmarks, spanning different triggers, geologies, and regions, highlight the need for methods that generalize beyond individual case studies and curated patches \citep{meena2023hr}.

Combining optical and SAR data leverages their complementary properties, optical imagery captures detailed surface characteristics, while SAR ensures consistent acquisition under all weather and illumination conditions. This fusion enhances the robustness and reliability of landslide detection. Previous work has demonstrated the benefits of optical-SAR integration for applications such as rainfall-induced landslide reconstruction, post-failure stability assessment, and susceptibility mapping \citep{zhang2025spatiotemporal,ghorbanzadeh2020application,phakdimek2023combination}.

The value of multi-source data fusion has been well established in various EO monitoring tasks. Author's previous studies have demonstrated the effectiveness of integrating heterogeneous data sources, including the fusion of Sentinel-2 imagery, digital elevation models (DEM), and NOAA climate data for algal bloom severity monitoring \citep{nasios2025ai}, as well as combining SAR-based AI predictions with AIS data for maritime monitoring within the SatShipAI platform \citep{nasios2025satshipai}. Building on this, present work extends to the domain of geohazard assessment by jointly leveraging Sentinel-1 and Sentinel-2 data for landslide detection. Current study originated from the author's participation in the Classification for Landslide Detection ML competition hosted on the \href{https://zindi.africa/competitions/classification-for-landslide-detection/}{Zindi Africa} platform, where the proposed approach achieved the second place. The experience provided a strong empirical basis for refining and systematically evaluating the current multimodal methodology, emphasizing robustness, methodological innovation, and operational applicability.

The integration of post-event optical imagery with pre- and post-event SAR observations exemplifies a well-conceived dataset design to manage complexities of real-world hazard detection. This multimodal configuration accommodates a range of operational scenarios by addressing varying data availability conditions across space and time. Framing the landslide detection task as a classification problem, rather than a pixel-level segmentation one, is particularly appropriate given the spatial resolution of Sentinel imagery  (\autoref{fig:samplesNbands}). The classification framework not only aligns with the intrinsic properties of the data but also capitalizes on broader spatial and contextual patterns while mitigating the effects of local noise. This design has provided a valuable experimental foundation for the development and assessment of advanced multimodal methodologies, facilitating systematic integration of heterogeneous data sources within a unified modeling framework. By enabling the use of state-of-the-art (SOTA) ML architectures and ensemble strategies, such an approach effectively leverages the complementary characteristics of optical and SAR data. The outcomes of this study further reinforce the advantages of multimodal fusion, demonstrating its potential to improve landslide detection accuracy and to advance geohazard monitoring capabilities.

Transformer-based architectures have rapidly become a cornerstone in EO and RS, offering superior capabilities for modeling complex spatial relationships and long-range dependencies that traditional convolutional networks often overlook. Initially introduced in natural language processing, the self-attention mechanism at the core of transformers has been successfully adapted to EO imagery, enabling models to capture global contextual features crucial for scene understanding and classification \citep{bazi2021vision,schiller2024forest,zhang2025robust}. Vision Transformers (ViTs) have demonstrated remarkable performance across diverse tasks, including land cover classification, hyperspectral analysis, and scene recognition, often surpassing convolutional neural networks in accuracy and generalization \citep{bazi2021vision,wang2022land}. Recent developments, such as the Multi-Instance Vision Transformer (MITformer), further enhanced transformers' ability to preserve key local features within complex spatial arrangements through multiple instance learning formulations \citep{sha2022mitformer}, while shape-aware architectures like ShapeFormer have advanced landslide detection by effectively modeling multi-scale and boundary-sensitive features in optical imagery \citep{lv2023shapeformer}. These innovations highlight transformers' growing dominance in EO classification tasks, offering powerful and flexible frameworks that seamlessly integrate global and local feature representations. As summarized by \cite{aleissaee2023transformers}, transformers are reshaping the landscape of RS analytics, providing a unifying architecture that is both data-efficient and highly adaptable across modalities such as Sentinel-1 SAR and Sentinel-2 optical data.

Ensemble learning has emerged as a powerful paradigm in RS, enabling the integration of diverse models to improve prediction accuracy, generalization, and robustness. By combining multiple learners, ensemble frameworks can better capture nonlinear relationships inherent in complex geospatial data \citep{zhang2022review,yang2023survey}. In the context of EO, ensemble methods have proven effective across a wide range of applications, from vegetation and urban surface mapping to hazard monitoring, by mitigating overfitting and leveraging complementary strengths among base models. Modern ensemble deep learning techniques, which combine neural networks with traditional ML classifiers, have further advanced predictive performance, providing enhanced adaptability to heterogeneous data sources and large-scale satellite datasets \citep{yang2023survey}. Such hybrid systems are increasingly relevant in operational scenarios where both model accuracy and computational efficiency are critical.

In landslide-related studies, ensemble learning has demonstrated notable improvements over single-model approaches, achieving higher accuracy and more stable generalization across varied terrain and sensor conditions. \cite{kalantar2020landslide} showed that integrating multiple algorithms, including Random Forest, Gradient Boosting, and Generalized Logistic Models, yielded superior performance for landslide susceptibility mapping compared to any individual classifier. Similarly, \cite{fang2021comparative} demonstrated that heterogeneous ensemble techniques such as stacking and blending outperformed individual convolutional and recurrent neural networks in susceptibility mapping, achieving more reliable spatial predictions. These findings underscore the ability of ensemble frameworks to effectively combine models with different learning dynamics, enhancing both detection precision and robustness. Comparable advances have also been observed beyond landslide applications, such as in urban impervious surface extraction, where ensemble boosting and soft-voting strategies integrating SAR and optical data have achieved accuracies exceeding 90\% \cite{ahmad2024novel}. Collectively, these results highlight ensemble learning as a cornerstone of modern RS analytics, offering a scalable and generalizable solution for complex environmental and hazard monitoring tasks.

This study introduces a multi-modal framework that integrates Sentinel-2 optical imagery with Sentinel-1 SAR data to enhance landslide detection performance. Unlike conventional change-detection approaches, this task relies solely on post-event Sentinel-2 observations, while both pre- and post-event SAR acquisitions are utilized, making the problem formulation distinct from classical bi-temporal analysis. The proposed framework combines deep neural networks based on Transformer architectures with Gradient Boosting Models (GBMs), leveraging both raw spectral bands and derived indices such as the NDVI to enhance feature representations. By allocating dedicated model instances to each data modality, the approach enables the extraction of modality-specific patterns while effectively fusing complementary information. Furthermore, this work emphasizes the role of ensemble learning in enhancing robustness and reliability across heterogeneous models. The proposed methodology achieved top performance in landslide detection, demonstrating high accuracy and robustness across diverse evaluation settings.

\section{Material and Methods}
\label{sec:Material and methods}

\subsection{Data}
\label{sec:Data}
The data preparation pipeline was designed to fully exploit the available satellite imagery and support robust model training. The provided datasets consisted of Sentinel-2 optical (4 bands), post-event Sentinel-1 SAR (4 bands), and pre- and post-event Sentinel-1 band difference (4). To enrich the input feature space, feature engineering was performed to derive six additional channels, including the optical indices NDVI and NDWI, along with four other spectral combinations. For the GBM models, patch-level summaries were generated using statistical descriptors, allowing for efficient learning with tree-based methods. Band-wise normalization was applied to ensure comparability across modalities, and data augmentation strategies were employed to improve generalization, mitigate class imbalance, and increase robustness to spatial variability.

\subsubsection{Raw Datasets}
\label{sec:Dataset}
The training and test datasets each comprise 12-channel image patches integrating multi-modal information from Sentinel-1 and Sentinel-2. The optical component consists of four bands, Red, Green, Blue, and Near-Infrared (NIR), while the SAR component includes eight channels derived from both ascending and descending orbits, containing VV and VH polarizations and their corresponding change-detection bands (computed as the difference between pre- and post-event backscatter). The full set of channels is as follows: Red (0), Green (1), Blue (2), NIR (3), Descending VV (4), Descending VH (5), Descending Diff VV (6), Descending Diff VH (7), Ascending VV (8), Ascending VH (9), Ascending Diff VV (10), and Ascending Diff VH (11). These channel indices correspond to the feature importance plots in \autoref{fig:FI_lgbm_xgb}. Typical sample images spanning all 12 bands are presented in 
\autoref{fig:samplesNbands}. No further metadata, such as acquisition dates, geographic locations, or satellite product specifications, were made available.

\begin{figure}[h]
    \centering
    \captionsetup{width=.44\linewidth}
    \subfigure[]{\includegraphics[trim={0 0 0 0},clip, width=0.22\textwidth]{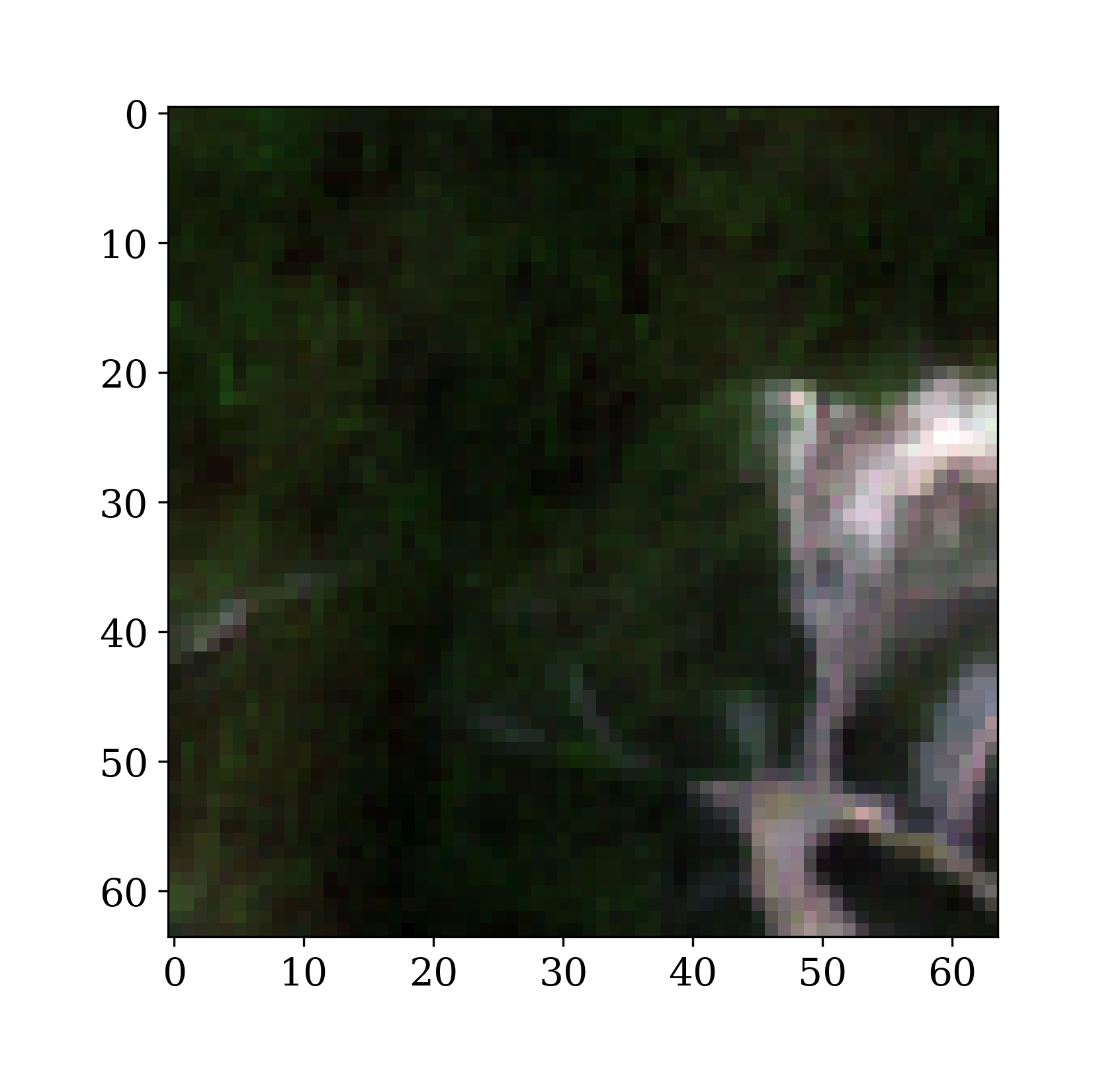}} 
    \subfigure[]{\includegraphics[trim={0 0 0 0},clip, width=0.22\textwidth]{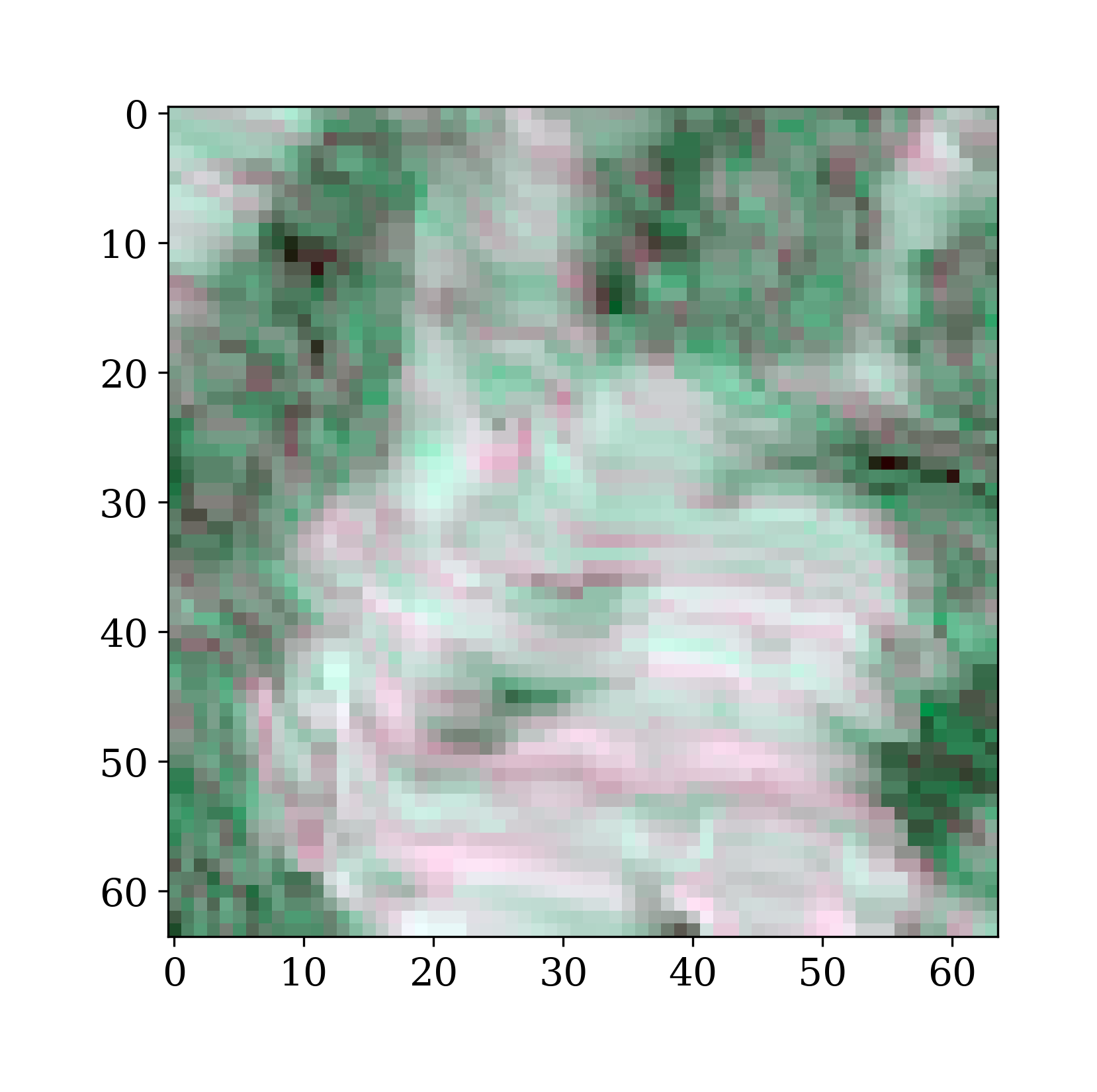}} 
    \caption{
    (a) Sentinel-2 RGB image, (b) Sentinel-1 pseudo-RGB SAR composite image}
    \label{fig:rgb_sar}
\end{figure}

Each patch measures 64×64 pixels at a spatial resolution of approximately 10 meters per pixel, covering roughly 640×640 meters on the ground. The training dataset consists of 7,147 patches with shape (7147, 64, 64, 12). The `Train.csv' file contains the ground truth value for every patch image. At the first column of the file there is the `ID', the patch image filename, and at the second the `label', indicating whether the patch represents a landslide (1) or a non-landslide (0) area. A visualization example is shown in \autoref{fig:rgb_sar}, where the left panel displays the RGB image and the right panel shows a pseudo-RGB SAR composite (Descending VV, VH, and their average). This is an imbalanced dataset, containing 5,892 non-landslide patches and 1,255 landslide patches ($\approx$ 17.5\%). This imbalance underscores the importance of robust validation strategies and the use of metrics suited to skewed datasets, such as the F1 score. The test dataset comprises 5,398 image patches and was employed to assess generalization performance, with results reported separately for the public and private leaderboard (LB), corresponding to a 30-70\% split, respectively.

\subsubsection{Feature Engineering}
\label{sec:feature_engineering}
In addition to the 12 channels provided in the original dataset, six supplementary bands (bands 12-17) were created to expand the feature space and support some of the models in \autoref{table:model_scores}. These extra channels were designed to capture vegetation status, water presence, and spectral relationships that could provide stronger indicators of landslide activity.

These bands included the two well-known indices, the NDVI of \autoref{eqn:ndvi}, which highlights vegetation health and density, and the Normalized Difference Water Index (NDWI), which emphasizes moisture and water features. Both are relevant in the context of landslides, since slope failures are often associated with sudden vegetation loss and altered moisture conditions. Additionally, four band-ratio features were constructed by combining pairs of the available optical channels (Red, Green, Blue, and NIR). To create the rest of the indices in places of NIR and Red bands of \autoref{eqn:ndvi} the following combinations were used: NIR-Green (NDWI), NIR-Blue, Blue-Green, Blue-Red and Green-Red. These ratios help to emphasize subtle spectral differences and interactions that may not be as easily detected in the original bands alone. By integrating these engineered features, the models had access to complementary information that improved their ability to separate landslide-affected areas from stable terrain.

\begin{equation}
\label{eqn:ndvi}
\text{NDVI} = \frac{NIR - Red}{NIR + Red + 10^{-10}}
\end{equation}

\subsubsection{Data Preprocessing}
\label{sec: Data Preprocessing}
Two primary preprocessing strategies were applied in the neural network experiments, while no preprocessing was performed for the LightGBM and XGBoost models. Specifically, standard scaling and robust scaling (using the 5th and 95th percentiles) were employed to normalize the input data and mitigate the influence of outliers (\autoref{table:model_scoresScaling}). To ensure stable and representative statistics, the mean, standard deviation, and percentile values for each band were computed using the combined training and test datasets, resulting in more reliable scaling parameters.

\subsubsection{Create Datasets for Gradient Boosting Models}
\label{sec:Csreate datasets}
For the GBMs (LightGBM and XGBoost), a feature-based dataset was constructed (1D), derived from the original image patches (2D). Instead of using raw pixel values directly, statistical descriptors were computed for each channel within a patch, transforming spatial information into compact numerical summaries. Specifically, seven statistics, minimum, mean, median, maximum, standard deviation, skewness, and kurtosis, were calculated independently for every channel of each patch.

With 18 available channel (12 raw + 6 calculated indices), this procedure produced a total of 126 (=7*18) features per patch, forming the input for the gradient boosting models. This approach allowed the models to capture both central tendencies and distributional properties of the spectral data, while reducing dimensionality compared to pixel-level training. By relying on aggregated features, GBMs were able to efficiently exploit spectral and textural differences between landslide-affected and unaffected areas.

\subsubsection{Augmentation}
\label{sec: Augmentation}
Augmentations included vertical flips, horizontal flips, and 90-degree rotations, each applied with a 50\% probability. During inference, test-time augmentation (TTA) was performed using all four possible flip combinations to improve prediction robustness. Both training and inference for the neural networks were conducted on images resized to 256 $\times$ 256 pixels, whereas the gradient boosting models operated on the original feature-derived data, requiring no image resizing and no augmentation.

\subsection{Evaluation Metric}
\label{sec:Metric}

To assess the performance of the proposed landslide detection models, the F1 score was used as the primary evaluation metric. The F1 score is particularly suitable for imbalanced classification problems, such as landslide mapping, where the number of landslide-affected cases (positive class) is typically much smaller than that of non-landslide cases (negative class). 

The F1 score is defined as the harmonic mean of precision ($P$) and recall ($R$):

\begin{equation}
\label{eqn:f1score}
F1 = 2 \cdot \frac{P \cdot R}{P + R},
\end{equation}

where precision and recall are given by:

\begin{equation}
\label{eqn:PR}
P = \frac{TP}{TP + FP}, \quad R = \frac{TP}{TP + FN}.
\end{equation}

Here, $TP$ denotes true positives (correctly detected landslide cases), $FP$ false positives (non-landslide cases misclassified as landslides), and $FN$ false negatives (landslide cases missed by the model).

Precision reflects the reliability of positive predictions, whereas recall captures the model's ability to identify actual positives \citep{goutte2005probabilistic}. The F1 score combines these two metrics into a single harmonic mean, balancing omission and commission errors. This makes it a more informative indicator of model quality than accuracy, particularly for imbalanced datasets, as it jointly accounts for false alarms and missed detections.

In this study, model training and threshold tuning were explicitly guided by F1 optimization, ensuring that the final predictions are robust to class imbalance and aligned with the evaluation criteria.

\subsection{Modeling}
\label{sec:Modelling}
\begin{figure}[h]
    \centering
    \captionsetup{width=.99\linewidth}
    \includegraphics[width=0.99\textwidth]{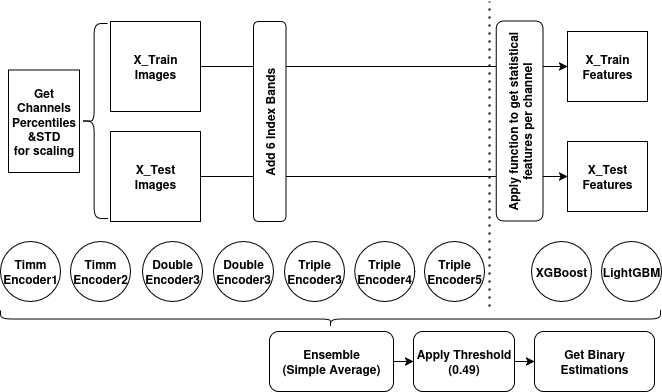}
    \caption{Flowchart}
    \label{fig:Landslide_class_flowchart}
\end{figure}

The proposed framework adopts a robust ensemble strategy, integrating nine classification models, including seven neural networks and two gradient boosting models. The overall methodology is summarized in \autoref{fig:Landslide_class_flowchart}. In addition to the original spectral bands, several derived spectral indices were incorporated into selected models (see \autoref{table:model_scores}). Model predictions were averaged to form ensemble probabilities which were subsequently binarized to maximize the F1 score. This approach increases both predictive performance and generalizability, yielding a robust and reliable solution for landslide detection.

\subsubsection{GBM Hyperparameters}
\label{sec:GBMhyper}
GBMs offer a wide range of tunable hyperparameters, which can be adjusted to optimize performance for a specific dataset and task. In this work, key parameters were carefully selected to balance convergence stability, generalization, and robustness to class imbalance. For the XGBoost model \citep{chen2015xgboost}, a low learning rate (0.02) combined with a large number of estimators (4000) ensured gradual and stable learning. Generalization was further improved through row subsampling (0.8) and column sampling by tree (0.8). Class imbalance was mitigated using a scale\_pos\_weight of 1.5, while L2 regularization (lambda = 1.2) reduced overfitting.

For the LightGBM model \citep{ke2017lightgbm}, a similar approach was adopted, with 24 leaves and a maximum tree depth of 12 allowing for moderately complex decision boundaries. A very small learning rate (0.005) paired with 12,000 boosting iterations promoted fine-grained updates. Subsampling (0.6) and column sampling (0.5) were employed to enhance model generalization, while class imbalance was again handled with a scale\_pos\_weight of 1.5. Additional regularization was provided by L2 (reg\_lambda = 0.3) to prevent overfitting.

Both GBMs followed very similar design philosophies and achieved comparable performance on the datasets (\autoref{table:model_scores}). Both applied higher weight to positive samples to address class imbalance, included L2 regularization to mitigate overfitting, and used subsampling to enhance generalization. However, LightGBM employed a smaller learning rate with three times more boosting rounds, allowing finer-grained updates and deeper trees capable of capturing more complex structures. 

\subsubsection{NN Architectures}
\label{sec:NNarchs}
For the neural network component, the PyTorch framework was utilized in conjunction with the timm library \citep{rw2019timm}. Pretrained timm models were either employed directly or integrated within custom-designed architectures. These custom configurations incorporated multiple instances of the timm encoder, with each instance assigned to a distinct subset of input channels, such as RGBN, SAR, SAR difference, or engineered optical indices, thereby enabling the network to learn modality-specific representations and effectively capture complementary features across data sources.

\autoref{table:encoders} lists the timm models used in the final ensemble, the broader architecture each belongs to, and the simplified encoder names adopted in this study. These shorter names are introduced for clarity, as the original model identifiers are lengthy and primarily reflect details of their implementation.

\begin{table}[h]
\caption{Encoders used in ensemble}
\label{table:encoders}
\centering
\begin{tabular}{l l l } 
 \hline
 Timm model name & Architecture & Encoder  \\ 
 \hline
maxvit\_rmlp\_tiny\_rw\_256.sw\_in1k & Maxvit & Enc1  \\ 
vit\_medium\_patch16\_reg4\_gap\_256.sbb\_in12k\_ft\_in1k & ViT &Enc2  \\ 
vit\_pwee\_patch16\_reg1\_gap\_256.sbb\_in1k & ViT & Enc3  \\ 
caformer\_s18.sail\_in22k\_ft\_in1k & CaFormer & Enc4  \\ 
vit\_medium\_patch16\_rope\_reg1\_gap\_256.sbb\_in1k & ViT & Enc5  \\ 
\end{tabular}
\end{table}

\begin{figure}[h]
    \centering
    \captionsetup{width=.99\linewidth}
    \includegraphics[width=0.99\textwidth]{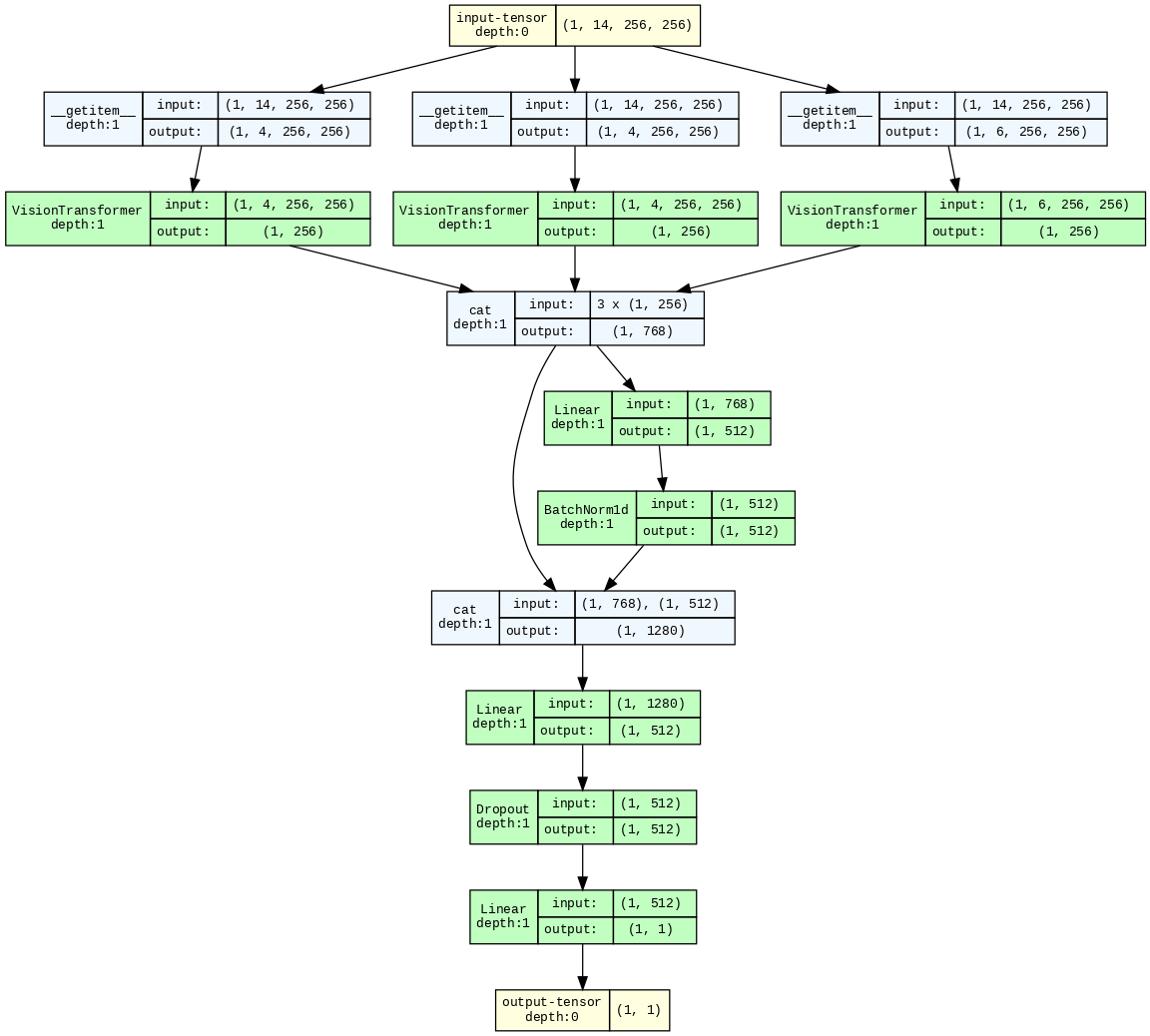}
    \caption{Model with 3 input modalities}
    \label{fig:model3inputs}
\end{figure}

\autoref{fig:model3inputs} depicts one of the models included in the final ensemble. This consists of three parallel instances of Enc3, each producing 256 output features. The first instance processes RGBN bands, the second processes the four SAR difference, and the third processes the six optical indices. The feature outputs from all three instances are concatenated and passed through a series of dense, batch normalization and dropout layers before reaching a single-logit output layer. This design allows each modality to learn modality-specific representations while also leveraging their combined information for improved predictive performance. The lightweight ResNet-like block that was integrated at the feature level, after processing the individual modalities, refined and aligned their representations. This residual design enhanced local feature propagation and cross-modal consistency, complementing the transformer's global attention with improved stability and representational depth \citep{he2016deep}.

Among the pretrained encoders that were used either as standalone or as base for a larger architecture, five distinct pretrained architectures were employed, with Enc3 been used in three different models. Overall, the final ensemble relies on three main transformer architectures: ViT \citep{dosovitskiy2020image}, Maxvit \citep{tu2022maxvit} and CaFormer \citep{yu2023metaformer}.

\subsubsection{Training Scheme}
\label{sec: Training Scheme}
A 5-fold cross-validation strategy was employed to ensure robust model training and reliable performance estimation. Derived out of fold (OOF) predictions not only provided an unbiased estimate of model performance but were also instrumental in tasks such as ensembling and calibration of classification thresholds. The approach reduces the risk of overfitting to a particular train/validation split and improves the generalizability of the final models.

All models were trained using the same 5-fold cross-validation split, ensuring that the resulting OOF predictions are consistent and free from leakage. To promote diversity within the ensemble and enhance its overall performance, the models were trained on different combinations of data modalities and scalings (see \autoref{table:model_scores}). This variation in input configurations encourages complementary error patterns, leading to more robust and accurate ensemble predictions.

Neural network training was performed using a custom cosine annealing learning rate schedule, with most models trained for 20 epochs using the AdamW optimizer. The loss function that was used is a simple average of BCEWithLogitsLoss and SmoothF1Loss \citep{benedict2021sigmoidf1}, allowing the model to balance probabilistic calibration with F1 score optimization.

\subsubsection{Calibration of Ensemble Outputs}
\label{sec: Final Predictions}
The final classification outputs were derived by averaging the probability estimates produced by all models included in the ensemble. OOF predictions were subsequently employed to guide the calibration process, enabling both refinement of the ensembling strategy and determination of the optimal decision threshold. A threshold value of 0.49 was identified as optimal, as it maximized the F1 score on the OOF data. This calibration step improved the robustness and reliability of the binary classification results. Notably, the proximity of the selected threshold to the default value of 0.5 reflects the well-calibrated nature of the ensemble's probability estimates and further substantiates the stability of the proposed methodology.

\subsection{Methodological Advancements in Multi-Modal Landslide Detection}
\label{sec:Innovation}
The proposed solution introduces several innovative components aimed at maximizing the accuracy and robustness of landslide detection from multi-source satellite imagery. A central contribution is the multi-encoder neural network architecture, in which the same pretrained timm encoder is instantiated multiple times, each instance dedicated to a distinct input modality (RGBN, SAR, SAR-difference, or derived index channels). Using a consistent encoder design across modalities ensures compatibility with training parameters such as optimization strategies and learning rate schedules while maintaining balanced representational capacity. This architecture also facilitates interaction between modalities at the feature level as each encoder first processes modality-specific information independently, and the resulting representations are then fused to capture both unique characteristics and cross-modal relationships. The fused features thus exploit the complementary strengths of optical precision and SAR's all-weather resilience, yielding more discriminative and robust inputs for the final classification layers.

Another important methodological contribution lies in the incorporation of derived spectral indices as a separate model input stream. In addition to the original spectral channels, derived indices, such as the NDVI and other vegetation- and moisture-related transformations, were integrated to enrich the input feature space. Treating these indices as a separate modality enables the network to learn specialized representations of vegetation dynamics, soil exposure, and surface moisture variability, all of which are critical indicators of landslide activity. This design choice enhances the model's capacity to capture subtle environmental changes that may not be fully represented in the raw optical or SAR data alone.

Taken together all design choices represent a comprehensive and innovative strategy for landslide detection. The approach integrates SOTA vision transformers encoders with custom multi-encoder architectures, combines heterogeneous model families (NNs and GBMs) within a unified ensemble learning technique, and leverages enriched feature sets derived from both optical and SAR modalities. By incorporating multiple data scaling strategies, a specialized loss function, and modality-specific model instances, the framework promotes complementary error patterns that strengthen ensemble performance. The methodology is further optimized through metric-driven threshold tuning, resulting in a solution that achieves SOTA performance while maintaining robustness, adaptability, and strong generalization to unseen data.

\section{Results and Discussion}
\label{sec:Results and discussion}

\autoref{table:model_scores} summarizes all models included in the ensemble, detailing their respective input data modalities, scaling strategies, and performance metrics on both OOF validation and competition LBs. The GBMs achieved competitive performance, providing a reliable and computationally efficient option for low-resource environments. All NN generally outperformed the GBMs, with the original optical bands (RGBN) being critical for achieving optimal accuracy. Notably, the NN trained solely on optical features achieved performance slightly superior to GBMs trained on both optical and SAR data. The best performance was obtained when NNs leveraged both optical and SAR inputs, capturing complementary information across modalities. Finally, ensembling all nine models enhanced overall performance while improving prediction confidence and robustness, resulting in a more reliable and trustworthy solution.

The overall F1 score as reported in result tables is a simple average of OOF and LB results, expressed as:

\begin{equation}
\label{eqn:overall}
Overall = 0.5 \times \text{OOF} + 0.5 \times (0.3 \times \text{Public LB} + 0.7 \times \text{Private LB})
\end{equation}

In \autoref{eqn:overall}, the factors 0.3 and 0.7 correspond to the relative proportions of public and private LB data, while the factor 0.5 assigns equal weight to training (OOF) and testing (LB) performance, providing a more balanced and fair evaluation metric.

\begin{table}[t]
\caption{Models, data modalities used, scaling and performance}
\label{table:model_scores}
\centering
    \addtolength{\leftskip} {-3cm}
    \addtolength{\rightskip}{-3cm}
\begin{tabular}{l  l  cccc l l l l l} 
 \hline
 Model & Encoder & RGBN & SAR & SAR diff & Indices & Scaling & OOF & AUC & pu/pr LB & Overall\\ 
 \hline
LightGBM & & \checkmark & \checkmark & \checkmark & \checkmark & None & 0.8740 & 0.9857 & 0.8792/0.8408 & 0.8632 \\ 
XGBoost  & & \checkmark & \checkmark & \checkmark &  \checkmark & None & 0.8708 & 0.9843 & 0.8786/0.8468 & 0.8636\\ 
Timm &  Enc1 &  \checkmark & & &  & Robust & 0.8744 & 0.9834 & 0.8854/0.8625 & 0.8719 \\ 
Timm &  Enc2 &  & \checkmark & \checkmark & \checkmark & Robust & 0.8496 & 0.9736 & 0.8475/0.8524& 0.8503 \\ 
combinedV2 & Enc3 & \checkmark & & \checkmark &  & Standard & 0.8976 & 0.9850 & 0.9148/0.8884 & 0.8970 \\ 
combinedV2 & Enc3 & \checkmark & & \checkmark &  & Robust & 0.8963 & 0.9871 & 0.9118/0.8908 & 0.8967\\ 
combinedV3 & Enc3 & \checkmark & & \checkmark & \checkmark & Standard & 0.9003 & 0.9862 & 0.9136/0.8828 & 0.8962\\ 
combinedV3 & Enc4 & \checkmark & & \checkmark & \checkmark & Standard & 0.8986 & 0.9872 & 0.9082/0.9001 & 0.9006\\ 
combinedV3c & Enc5 & \checkmark & \checkmark & \checkmark &  & Standard & 0.8943 & 0.9894 & 0.9100/0.8956 & 0.8971\\ 

 \hline
Ensemble & & & &  & & & 0.9216 & 0.9928 & 0.9409/0.9058 & 0.9190 \\ 

\end{tabular}

\end{table}

Because the F1 score may not be intuitively interpretable in all cases, an alternative perspective on the OOF predictions is presented in \autoref{table:conf_matrix}. This confusion matrix shows that most predictions correspond to True Negatives (correctly identified non-landslide areas) and True Positives (correctly detected landslide areas), while the number of False Positives (incorrectly classified as landslides) and False Negatives (missed landslides) remain low, indicating overall reliability and balanced performance. Calculating the precision and recall (\autoref{eqn:PR}), the ensemble achieves approximately 95\% precision and 90\% recall. This implies that when the model predicts a landslide, it is correct 95\% of the time, while still identifying 90\% of all actual landslide occurrences. To further evaluate the OOF predictions, AUC scores \citep{fawcett2006introduction} were included in metrics of \autoref{table:model_scores}. Individual models achieved AUC values between 0.974 and 0.990, while the ensemble reached 0.993. These results are further illustrated by the ROC curves in \autoref{fig:ROCcurve}.

\begin{table}[ht!]
\caption{OOF Confusion matrix}
\label{table:conf_matrix}
\centering
\begin{tabular}{l  l  l  l } 
 \hline
 TN & FP & FN & TP \\ 
 \hline
5832 & 60 & 129 & 1126 \\ 

\end{tabular}

\end{table}

The proportion of predicted positive cases (True Positives + False Positives) in the training/validation data was approximately 16.6\%, closely matching the true positive rate of 17.5\%. For the unknown label of the test dataset, positive predictions accounted for about 11.4\% of all samples, indicating a lower prevalence of landslide events compared to the training set. This discrepancy suggests that the training data may have included a higher concentration of positive cases, which could have facilitated model learning and potentially contributed to improved performance on the training distribution.

The performance of the proposed methodology on the competition's private LB is presented in \autoref{table:comp_results}. Most submissions significantly outperformed the benchmark, with the top-ranked solutions achieving particularly high scores. The proposed approach achieved results comparable to the leading entry on this dataset, demonstrating its SOTA performance.

\begin{table}[ht!]
\caption{Top ML Competition Results}
\label{table:comp_results}
\centering
\begin{tabular}{l  l  l   } 
 \hline
 Rank & User & Score \\ 
 \hline
1 & 3B & 0.906892382 \\ 
\textbf{2 }& \textbf{ouranos} & \textbf{0.905797101} \\ 
3 & AhmedTambal & 0.902843601 \\ 
 \hline
Benchmark &&  0.576652601 \\ 
	 \hline
 \multicolumn{3}{l}{\footnotesize{ *Author's rank in bold }} \\
\end{tabular}

\end{table}

\subsection{Impact of Multi-Encoder Architectures on Model Performance}
\label{multiencoderimpact}

\begin{table}[t]
\caption{Multi-encoder vs Single-Encoder}
\label{table:model_scoresME}
\centering
    \addtolength{\leftskip} {-3cm}
    \addtolength{\rightskip}{-3cm}
\begin{tabular}{l  l  cccc l l l l} 
 \hline
 Model & Encoder & RGBN & SAR & SAR diff & Indices & Scaling & OOF & pu/pr LB & Overall\\ 
 \hline
 
combinedV4 & Enc3 & \checkmark & \checkmark & \checkmark & \checkmark & Standard & 0.9026 & 0.914/0.8958 & 0.9017 \\ 
timm & Enc3 & \checkmark & \checkmark & \checkmark & \checkmark & Standard & 0.8495 & 0.8806/0.8511 & 0.8547 \\ 
\hline
combinedV3 & Enc4 & \checkmark & & \checkmark & \checkmark & Standard & 0.8996 & 0.9082/0.9001 & 0.9011\\
timm & Enc4 & \checkmark & & \checkmark & \checkmark & Standard & 0.836 & 0.8762/0.8309 & 0.8402\\


\end{tabular}

\end{table}


Preliminary experiments demonstrated that aggregating all input bands into a single feature stack, rather than assigning each data modality to a dedicated model instance, led to a marked decline in performance. As shown in \autoref{table:model_scoresME}, experiments that processed modalities separately consistently outperformed those treating them jointly. This finding underscores the importance of allowing optical and SAR inputs to be modeled independently, enabling the network to learn modality-specific representations prior to feature fusion, which ultimately enhances both feature extraction and predictive accuracy. It is also noteworthy that the combinedV4 model with Enc3 encoder achieved the highest individual score, yet was excluded from the final ensemble because its inclusion did not improve the overall ensemble performance. This highlights a key consideration in ensemble design, that models with superior standalone performance do not always contribute positively to collective outcomes, and systematic evaluation of model combinations remains essential.

\subsection{Advantages of Vision Transformers over Conventional Encoders}
\label{ViTadvantages}

Exploratory experiments using architectures other than transformers yielded inferior results. Specifically, models from the EfficientNet, EdgeNeXt, and ConvNeXt architectures consistently underperformed compared to transformer-based approaches (\autoref{table:model_scoresRGBN}). This performance gap can be attributed to the inherent advantages of transformers, particularly their ability to capture long-range spatial dependencies and model global context across the entire image patch through self-attention mechanisms. These properties are crucial for landslide detection, where identifying subtle spatial patterns and context, such as changes in terrain structure or vegetation cover, often requires integrating information from non-local regions. Consequently, transformer architectures appear to be particularly well-suited for this task, offering a stronger representation capacity than convolution-based models.

\subsection{Impact of Model Pretraining on Robustness and Performance}
\label{pretrainingimportance}
The GBMs exhibited slightly better performance on the OOF data than on the private LB, implying mild overfitting to the training set. In contrast, the NNs, benefiting from ImageNet pretraining, converged faster than if were randomly initialized and, more importantly, demonstrated strong robustness, achieving nearly identical results across training and test datasets. This robustness was most evident when combining optical and SAR modalities. Interestingly, the model trained solely on optical data performed marginally better on the public LB but slightly lower on the private LB compared to its OOF score.

\subsection{Ensembling Learning and Threshold Considerations}
\label{ensembleNthreshold}
The final predictions were generated through an ensemble of nine models, combining seven neural networks and two gradient boosting models. This ensemble approach leveraged the complementary strengths of diverse architectures and data modalities, enhancing overall predictive accuracy and robustness beyond what individual models could achieve. The averaging of model outputs effectively mitigates individual biases and stabilizes predictions, providing consistent performance across both OOF data and LB evaluations.

\begin{figure}[h]
    \centering
    \subfigure[Ensemble]{\includegraphics[width=0.49\textwidth]{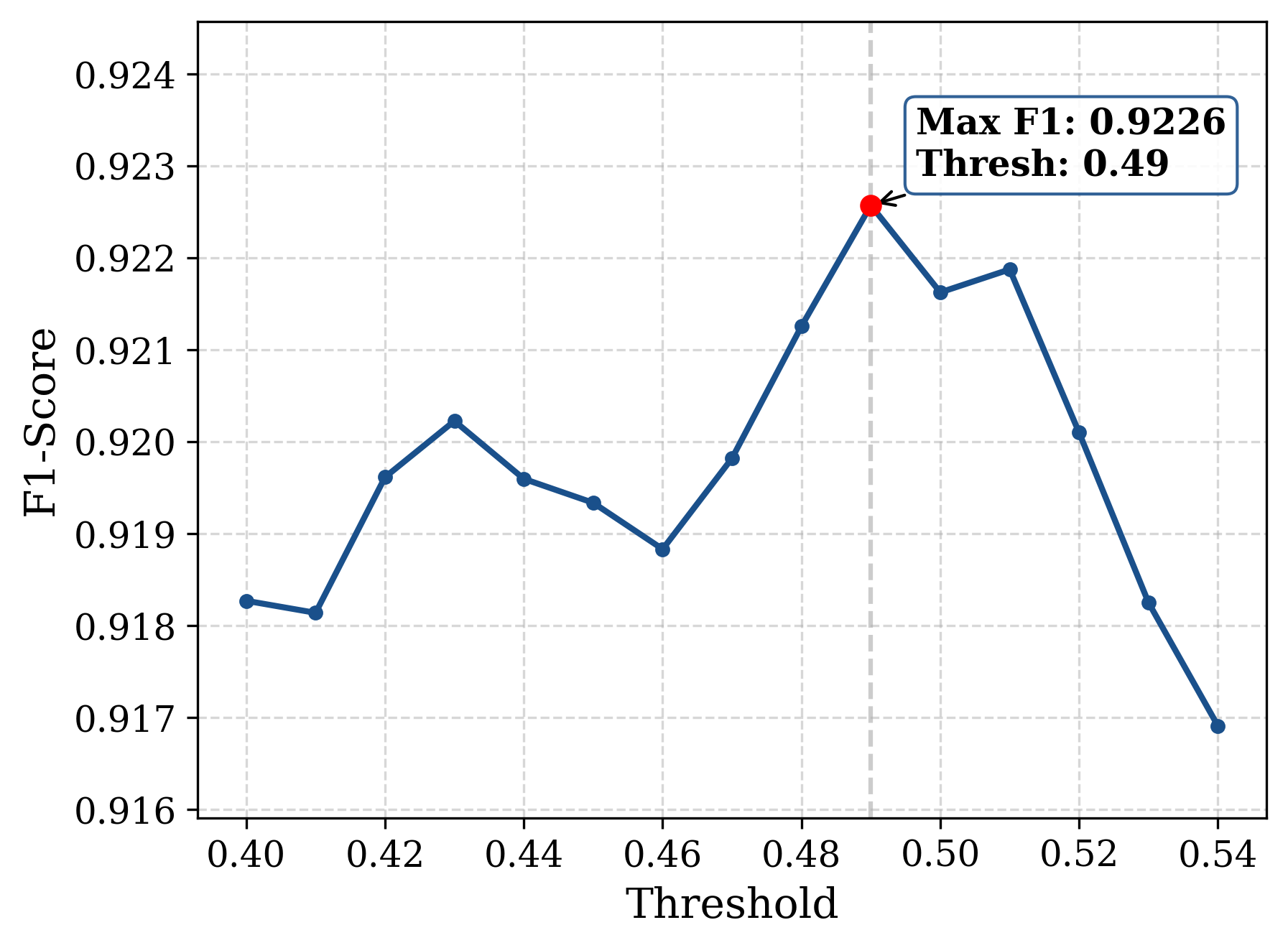}} 
    \subfigure[5 best performing NNs]{\includegraphics[width=0.49\textwidth]
    {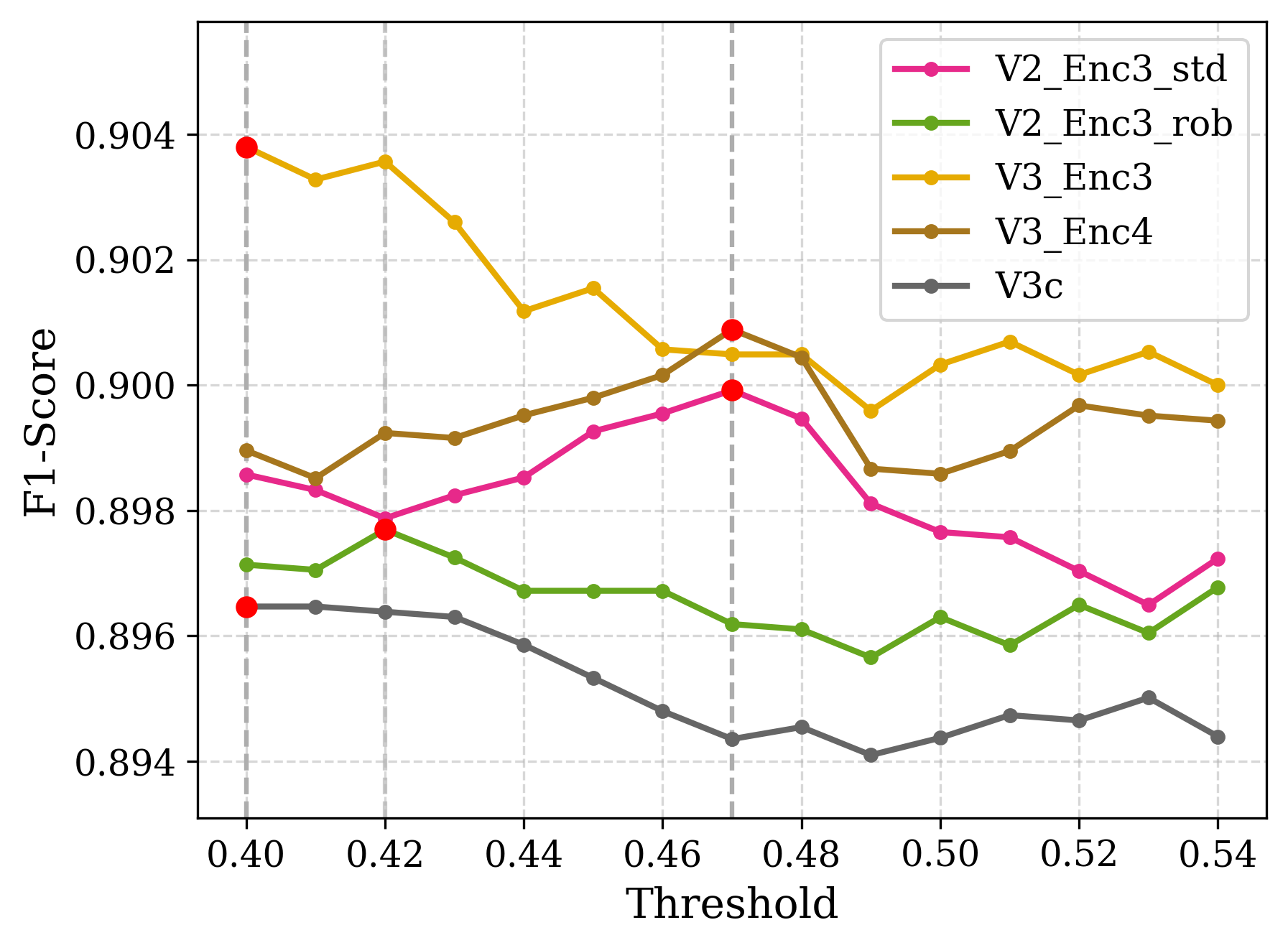}}     
    \caption{OOF score per Threshold}
    \label{fig:OOF_score_per_Threshold}
\end{figure}

While the ensemble inherently improves reliability, careful calibration of the probability to class threshold further refines binary classification performance. An optimal threshold of 0.49 was established based on OOF predictions (\autoref{fig:OOF_score_per_Threshold}), slightly below the default 0.5 and higher than the thresholds of individual models, achieving a balanced trade-off between precision and recall. However, the primary driver of the model's strong performance remains the diversity and combination of multiple high-performing models within the ensemble.

\subsection{LightGBM Ablation and Feature Importance Analysis}
\label{LGBablation}

To better quantify the contribution of individual data modalities to overall performance, ablation studies were conducted using the LightGBM model, chosen for its efficiency. In these experiments, model parameters were kept identical to the main training configuration. In the main study, each experiment systematically excluded one or two of the four modalities in each run. As shown in \autoref{table:model_scoresAS}, all reduced-modality configurations resulted in performance degradation, with models relying solely on either Sentinel-1 or Sentinel-2 data exhibiting the most pronounced decline. The poorest performance was observed when using only Sentinel-1 inputs (post-event SAR and SAR-difference bands). Notably, the derived optical indices proved to be particularly valuable, outperforming the experiment that used the raw optical bands, highlighting their importance for robust landslide detection.

\begin{table}[h]
\caption{LightGBM ablation study - Modalities included}
\label{table:model_scoresAS}
\centering
    \addtolength{\leftskip} {-3cm}
    \addtolength{\rightskip}{-3cm}
\begin{tabular}{  cccc l l l} 
 \hline
  RGBN & SAR & SAR diff & Indices &  OOF & pu/pr LB & Overall\\ 
 \hline
 \checkmark & \checkmark & \checkmark &  & 0.8551  & 0.8494/0.818 & 0.8413 \\ 
 \checkmark &  & \checkmark & \checkmark  & 0.8648  & 0.8759/0.8267 & 0.8531 \\ 
 \checkmark & \checkmark &  & \checkmark  & 0.8507  & 0.8516/0.8216 & 0.8407 \\ 
  & \checkmark & \checkmark & \checkmark  & 0.8629  & 0.8613/0.8335 & 0.8524 \\ 
  & \checkmark & \checkmark &  & 0.7692  & 0.7592/0.7414 & 0.758 \\ 
  \checkmark &  &  & \checkmark & 0.7978  & 0.8031/0.7861 & 0.7945 \\ 


\end{tabular}

\end{table}

A further ablation analysis was conducted to examine the suitability of the statistical descriptors used to derive features from the image data for the GBM models was examined. The results are summarized in \autoref{table:lightgbm_rem_stat}. In each experiment, one statistic was removed from the feature set, reducing the total number of features from 126 (18 channels × 7 statistics) to 108 (18 × 6). As expected, removing the `standard deviation' resulted in a noticeable performance decline, as this statistic effectively captures surface disturbance. A smaller drop occurred when the `minimum' statistic was removed, whereas excluding `skewness' led to a slight improvement in performance.

\begin{table}[h]
\caption{LightGBM ablation study - Remove Statistic}
\label{table:lightgbm_rem_stat}
\centering
    \addtolength{\leftskip} {-3cm}
    \addtolength{\rightskip}{-3cm}
\begin{tabular}{  l l l l} 
 \hline
  Statistic & OOF & pu/pr LB & Overall\\ 
 \hline
  Mean & 0.8705  & 0.8786/0.8481 & 0.8639 \\ 
  Median & 0.8697  & 0.8889/0.8412 & 0.8626 \\ 
  Min & 0.8695  & 0.8786/0.8398 & 0.8604 \\ 
  Max & 0.8733  & 0.8780/0.8395 & 0.8621 \\ 
  StD & 0.8621  & 0.8706/0.8285 & 0.8516 \\ 
  Skew & 0.8718  & 0.8862/0.8464 & 0.8651 \\ 
  Kurtosis & 0.8724  & 0.8689/0.8434 & 0.8617 \\ 

\end{tabular}
\end{table}

An additional ablation study evaluated the contribution of the derived indices used as inputs to the GBM models (see \autoref{table:lightgbm_rem_index}). In each experiment, one index was removed from the feature set, reducing the number of features from 126 (18 channels × 7 statistics) to 119 (17 × 7). Overall, the results showed only minor deviations from the full feature set. Notably, removing NDVI and NDWI had only a marginal impact on performance, while excluding the BlueGreen index slightly improved results and removing the GreenRed index led to a small performance decline.

\begin{table}[h]
\caption{LightGBM ablation study - Remove Index}
\label{table:lightgbm_rem_index}
\centering
    \addtolength{\leftskip} {-3cm}
    \addtolength{\rightskip}{-3cm}
\begin{tabular}{  l l l l} 
 \hline
  Index & OOF & pu/pr LB & Overall\\ 
 \hline
  NDVI & 0.8722  & 0.8818/0.8408 & 0.8627 \\ 
  NDWI & 0.8727  & 0.8792/0.8428 & 0.8632 \\ 
  NIRBlue & 0.8716  & 0.8780/0.8418 & 0.8621 \\ 
  BlueGreen & 0.8712  & 0.8819/0.8513 & 0.8658 \\ 
  BlueRed & 0.8732  & 0.8765/0.8467 & 0.8644 \\ 
  GreenRed & 0.8711  & 0.8710/0.8420 & 0.8609 \\ 
\end{tabular}
\end{table}

\begin{figure}[h]
    \centering
    \subfigure[LightGBM]{\includegraphics[trim={0 0 0 0},clip, width=0.49\textwidth]{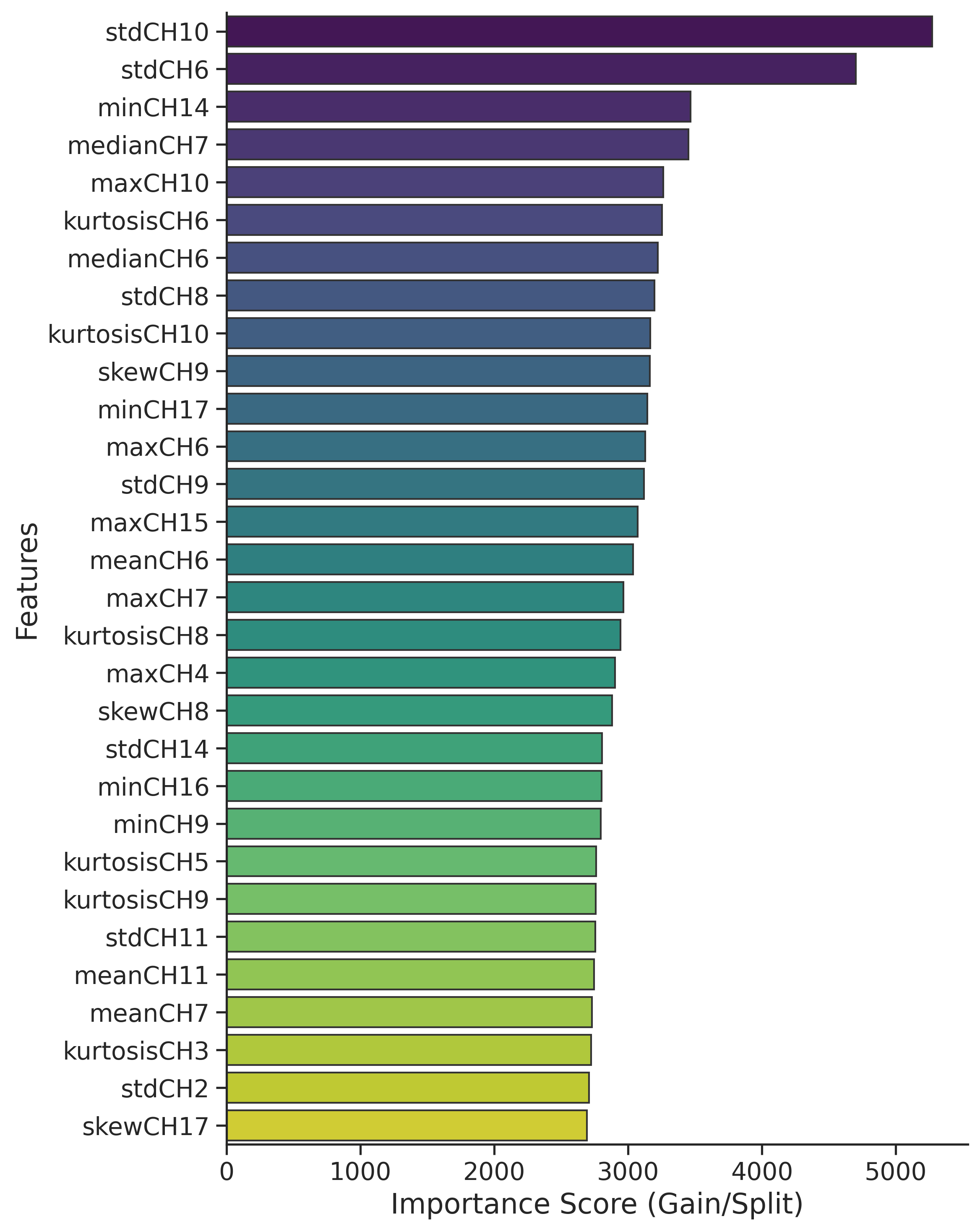}} 
    \subfigure[XGBoost]{\includegraphics[trim={0 0 0 0},clip, width=0.49\textwidth]{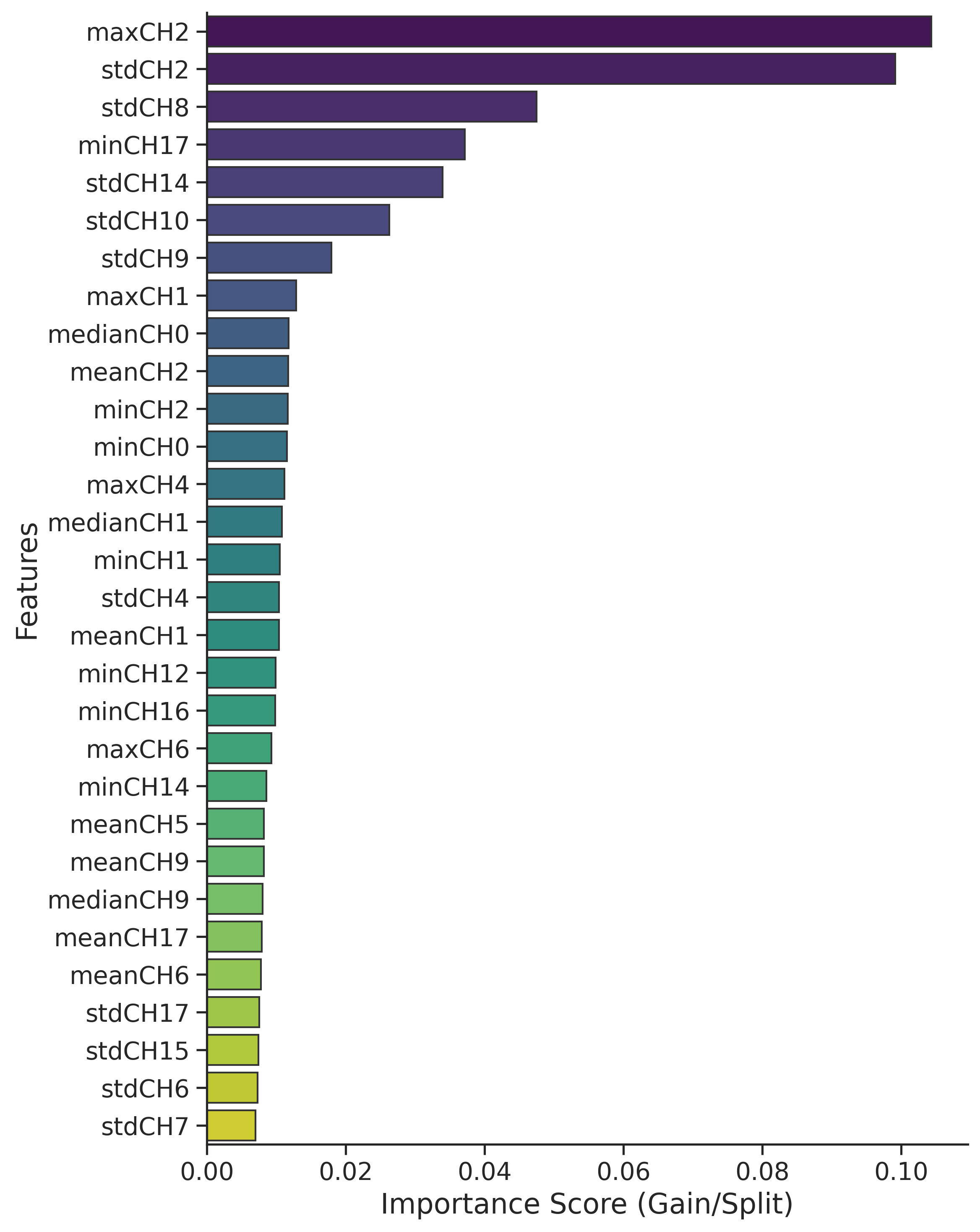}} 
    \caption{Feature Importance }
    \label{fig:FI_lgbm_xgb}
\end{figure}

To enhance interpretability and transparency, feature importance analysis was performed for both GBMs, revealing the relative influence of different input variables. \autoref{fig:FI_lgbm_xgb} presents the top 30 features, with results from LightGBM shown on the left and XGBoost on the right. Features with channel numbers greater than 11 correspond to derived indices. For LightGBM, the standard deviation of the \textit{Ascending Diff VV} and \textit{Descending Diff VV} band was identified as the most influential predictors, while in XGBoost the standard deviation and the maximum of the \textit{Blue} band dominated. Moreover, while LightGBM's top two features are derived from standard deviation across two channels, five of the top seven features in XGBoost are based on standard deviation. This is consistent with the findings of the ablation study, indicating that variability-related features within individual bands play a particularly important role in the decision-making process of the tree-based models.

Interpretability in neural networks is inherently more challenging, however, experimental evidence from performance comparisons revealed that the optical bands (RGBN) provided the strongest contributions to landslide detection. Importantly, the inclusion of SAR data consistently enhanced overall performance, demonstrating its complementary value and underscoring the benefit of multi-modal integration.

Taken together, these insights offer a clearer understanding of the ensemble's internal behavior, optical inputs serve as the primary discriminative features, while SAR data strengthens resilience and generalization. Such transparency not only aids in the validation of the modeling approach but also supports trustworthiness and accountability in potential real-world applications.

\subsection{Comparison with Recent State-of-the-Art Methods}
\label{comparisonstudies}
As no prior studies have been conducted on this specific dataset, a review of related work was undertaken. \cite{wang2022change} investigated change detection-based co-seismic landslide mapping using extended morphological profiles and an ensemble strategy, applying pre- and post-event Sentinel-2 images across three regions (Jiuzhaigou and Mainling in China, and Nippes in Haiti). Their reported average F1-score was 0.8356 (0.7841, 0.8450, and 0.8776, respectively). In another study, \cite{ren2024enhancing} explored the detection of old landslides in the Three Gorges Reservoir Area through a deep learning framework that incorporated Sentinel-2 imagery and topographical features, achieving an F1-score of 0.9105. While these studies were conducted on different datasets, the methodology presented in this work demonstrates superior performance, achieving an overall F1-score of 0.919 (\autoref{table:model_scores}). Its strong results in both the ML competition and in comparison with these SOTA methods further underscore the robustness and effectiveness of the proposed approach.

\subsection{Computational Efficiency and Inference Time}
\label{inferencetime}
All experiments were conducted on Google Colab Pro using a T4 GPU for NN training and inference, allowing for a direct comparison of execution speeds. \autoref{table:train_infer} show the time needed for training 1 epoch (in min:sec) and the time needed for inference the validation fold (without TTA) in the 5-fold validation setting used. As expected, both training and inference times scale with model size. Increasing the number of encoders within the model to process different data modalities in parallel resulted in proportionally longer runtime. The most commonly used, and relatively lightweight encoder, Enc3, required approximately 2 minutes for one training epoch and 10 seconds for inference when instantiated twice. Instantiating the same encoder three times within the model increased training time to 3 minutes 10 seconds and inference time to 16 seconds, while four instances required 4 minutes 10 seconds and 21 seconds, respectively. The other encoders used in the final ensemble were larger than Enc3, leading to approximately double the training and inference times. For the GBMs, training was performed on a standard CPU instance, with training taking only a few minutes and inference completing within a couple of seconds.

\begin{table}[ht!]
\caption{Execution Speeds Comparison}
\label{table:train_infer}
\centering
\begin{tabular}{l  ccc  } 
 \hline
 Encoder & Instances & Train 1 epoch & Inference\\ 
 \hline
Enc1 & 1 & 2:39 & 0:12 \\
Enc2 & 1 & 2:12 & 0:12 \\ 
Enc3 & 2 & 2:00 & 0:10 \\ 
Enc3 & 3 & 3:10 & 0:16 \\ 
Enc3 & 4 & 4:10 & 0:21 \\ 
Enc4 & 3 & 6:57 & 0:28 \\  
Enc5 & 3 & 6:36 & 0:32 \\ 

\end{tabular}

\end{table}

\subsection{Operational Considerations}
\label{sec:Practicality}

The proposed framework was designed with a strong emphasis on operational adaptability, ensuring its applicability under varying computational and data availability conditions. The performance of individual models, as well as their combinations, are summarized in \autoref{table:perf_pract}, encompassing both the models included in the final ensemble and some assessed during the experimental phase. This comparative evaluation provides insights into a range of potential deployment scenarios. For instance, the framework can be tailored to resource-limited environments where GPU access is restricted, or to cases where only a subset of the available modalities, such as Sentinel-1 SAR or Sentinel-2 optical imagery, can be utilized. Such flexibility underscores the capacity of the methodology to maintain robust performance across diverse operational constraints and application contexts.

\begin{table}[h]
\caption{Performance Comparison under Diverse Operational Scenarios}
\label{table:perf_pract}
\centering
\begin{tabular}{l  l l l  } 
 \hline
 Model & Encoder &  Note & F1 Score \\ 
 \hline
Timm &  Enc1 &  Optical Only & 0.8719 \\  
Timm &  Enc2 &  SAR Only & 0.806 \\ 
combinedV2 & Enc3 & Fastest NN & 0.897 \\ 
combinedV4 & Enc3 & All 4 modalities & 0.9019 \\ 
combinedV3 & Enc4 & Best Single used & 0.9011 \\ 
\hline
 Ensemble 2 GBMs  &  & CPU & 0.8666 \\ 
Ensemble 7 NN & & GPU & 0.9127 \\ 
Ensemble 9 & & GPU & 0.919 \\ 
 \hline
\multicolumn{4}{l}{\footnotesize{ * F1 Score is the Overall score }} \\

\end{tabular}

\end{table}

GBMs demonstrated competitive standalone performance and present clear advantages in settings constrained by computational resources. Unlike neural network-based approaches, GBMs do not depend on GPUs or other specialized hardware accelerators, which enables their deployment in low-resource or time-critical environments. Their ability to deliver rapid, reliable predictions with relatively low computational cost makes them well suited for large-scale, continuous monitoring, especially in settings where timely decision-making and operational efficiency are critical.

Sentinel-2 data were found to be more critical than Sentinel-1 data for achieving optimal performance. Experiments using only SAR data yielded lower performance emphasizing the benefit of incorporating optical information. 

This can be attributed to the rich spectral information provided by Sentinel-2, which captures variations in vegetation, soil exposure, and surface reflectance that are commonly associated with landslide scars and disturbed terrain. In particular, spectral bands enable the detection of vegetation loss and soil exposure, both strong indicators of landslide activity. In contrast, although Sentinel-1 SAR data provide valuable structural and moisture-related information and are robust to cloud cover and illumination conditions, their backscatter signals are often more difficult to interpret and may be influenced by terrain geometry, surface roughness, and speckle noise. As a result, SAR-only inputs may capture fewer discriminative features, whereas the inclusion of optical data significantly enhances the model's ability to distinguish landslide-affected areas. Though a SAR-only approach remains a viable option when optical imagery is unavailable. The optical-only neural network significantly outperformed the SAR-only approach, although both modalities can be valuable depending on the application. The ability of SAR to operate independently of cloud cover and during nighttime remains a key advantage in specific scenarios. Neural networks that fuse both optical and SAR data consistently delivered the highest accuracy for landslide detection. Among the models used in the ensemble, the best-performing single model was Enc4 (Caformer), which leveraged optical data, SAR differences, and optical indices. Interestingly, although not included in the final ensemble, the faster Enc3 (ViT-pwee), which utilized all four data modalities, achieved slightly higher performance.

All neural network encoders employed in this study are of tiny, small, or medium size. Although instantiating multiple encoders within a single model increases its overall complexity, the approach proved beneficial for performance. Among the tested configurations, the Enc3 architecture emerged as the most effective, being used extensively during both experimentation and in the final ensemble. Notably, the model variant that instantiated Enc3 twice, processing optical and SAR-difference modalities, demonstrated competitive accuracy while maintaining relatively fast inference. While GPU acceleration is recommended for training, deployment could be performed on CPU-only systems. When tested on a Google Colab Pro CPU environment, inference over 1,430 validation images required approximately 3 minutes, corresponding to $\approx$ 0.126 seconds per image, or nearly 8 images per second.

Overall, the pipeline is modular and scalable, with clearly defined components that can be configured to meet diverse deployment requirements. The use of widely adopted libraries and structured workflows facilitates integration into existing systems, underscoring the practicality of the approach for both research and operational contexts.

\subsection{Limitations}
\label{sec:Limitations}
Despite the strong performance of the proposed framework, several limitations should be acknowledged. Although ensemble learning contributes to improved predictive accuracy, it also increases model complexity and training time compared to single-model approaches, which may affect scalability in large-scale or time-sensitive applications. In addition, the dataset used in this study originates from a ML competition and was provided without accompanying metadata, such as acquisition dates, geographic locations, or detailed satellite product specifications, limiting deeper analysis of temporal and regional factors. Furthermore, the dataset construction process and sampling strategy remain unknown. While it offers a useful benchmark for model development, further evaluation on larger satellite scenes and operational datasets would be needed to better assess the robustness and real-world applicability of the proposed approach.

\subsection{Transferability and Methodological Extensions}
\label{sec:Perspectives}

The proposed methodology was developed with adaptability as a central principle, offering strong potential for extension beyond landslide detection to a wide range of disaster monitoring and environmental change applications. By combining heterogeneous model families, GBMs and NNs, the framework enables strategic model selection depending on computational resources and operational constraints. Its modular design allows seamless reconfiguration to operate with Sentinel-1 SAR or Sentinel-2 optical data independently, ensuring flexibility when modality availability is limited.

Looking forward, several promising directions may further enhance performance and broaden applicability. Incorporating additional Sentinel-2 bands, particularly the Shortwave Infrared (SWIR) channels, could significantly improve sensitivity to soil moisture, vegetation stress, and surface composition, all of which are critical for landslide and hazard-related assessments \citep{lu2021co,ren2024enhancing}. Expanding the spectral range within the current framework of only post-event Sentinel-2 imagery, could therefore strengthen model discrimination and overall robustness.

In addition, alternative data configurations offer complementary opportunities for improvement. Leveraging interferometric SAR (InSAR) products derived from pre- and post-event Sentinel-1 acquisitions could significant improve landslide detection. Likewise, incorporating pre- and post-event Sentinel-2 imagery would introduce valuable temporal context, enabling more explicit change detection analyses and supporting synergistic fusion strategies between optical and radar modalities.

These extensions build naturally on the framework's modular design, which already supports diverse input configurations and feature sets. Expanding in this direction would strengthen the methodology's applicability to other domains, such as flood extent mapping, earthquake damage assessment, wildfire burn scar detection, and broader environmental monitoring. By combining multi-modal inputs, lightweight transformer architectures, and ensemble learning strategies, the framework provides a robust foundation for future research, with the potential to scale across regions, sensor platforms, and a wide range of hazard contexts.

\section{Conclusions}
\label{sec:Conclusions}
This study presented a multi-modal framework for landslide detection that leverages the complementary information provided by post-event Sentinel-2 optical imagery and pre/post-event Sentinel-1 SAR data. The proposed approach demonstrated the overall superiority of the integrated methodology, particularly through the use of multi-encoder vision transformers in which each data modality is processed by a dedicated encoder. This architectural design was found to be especially effective, confirming the strong suitability of transformer-based encoders for complex multi-modal geospatial classification tasks.

Beyond the individual model performance, the study highlights the power of ensemble learning, where a combination of seven neural networks and two gradient boosting models (GBMs) delivered state-of-the-art classification results. The inclusion of GBMs (LightGBM and XGBoost) provided resource-efficient alternatives that can be deployed without specialized accelerator, ensuring scalability to environments with varying computational resources.

Feature engineering further enhanced performance, with the incorporation of optical indices such as NDVI and NDWI adding value by enriching the representation of terrain and land cover properties. The selected ensemble strategy, coupled with calibrated threshold, achieved a strong balance between precision and recall, yielding robust and generalizable results despite data imbalance.

The modular and flexible design of the proposed methodology allows it to perform effectively across a wide range of operational scenarios, accommodating optical-only, SAR-only, or combined data inputs. This adaptability also facilitates transfer to related environmental monitoring tasks, such as flood extent mapping, earthquake damage assessment, and wildfire detection, indicating its versatility and robustness across different deployment contexts.

Overall, the results demonstrate that multi-encoder vision transformers, when combined with ensemble learning, advanced feature engineering, and heterogeneous model families, provide a powerful and versatile solution for geohazard detection. The methodology not only sets a new benchmark for landslide classification but also establishes a scalable, generalizable, and operationally viable foundation for broader applications in disaster monitoring and environmental change detection.

\section*{Declaration of Competing Interest}
The author declares that has no known competing financial interests or personal relationships that could have appeared to influence the work reported in this paper.

\section*{Acknowledgment}
\label{sec:Acknowledgment}
I sincerely thank the anonymous reviewers for their insightful comments and suggestions, which have significantly contributed to improving the quality of this manuscript.

\section*{Data availability}
\label{sec:Data availability}
The data can be downloaded from the competition's website after logging in, at: \href{https://zindi.africa/competitions/classification-for-landslide-detection/data}{https://zindi.africa/competitions/classification-for-landslide-detection/data}.

\section*{Code availability}
\label{sec:Code availability} The complete running code as described in present research is available on \url{https://github.com/IoannisNasios/sentinel-landslide-cls}.

\bibliographystyle{cas-model2-names}

\bibliography{refs}

\newpage
\appendix 

\section{Typical sample image patches and model ROC curves}
\label{sec:Appendix A}

\begin{figure}[h]
    \centering
    \captionsetup{width=.5\linewidth}
    \includegraphics[width=0.99\textwidth]{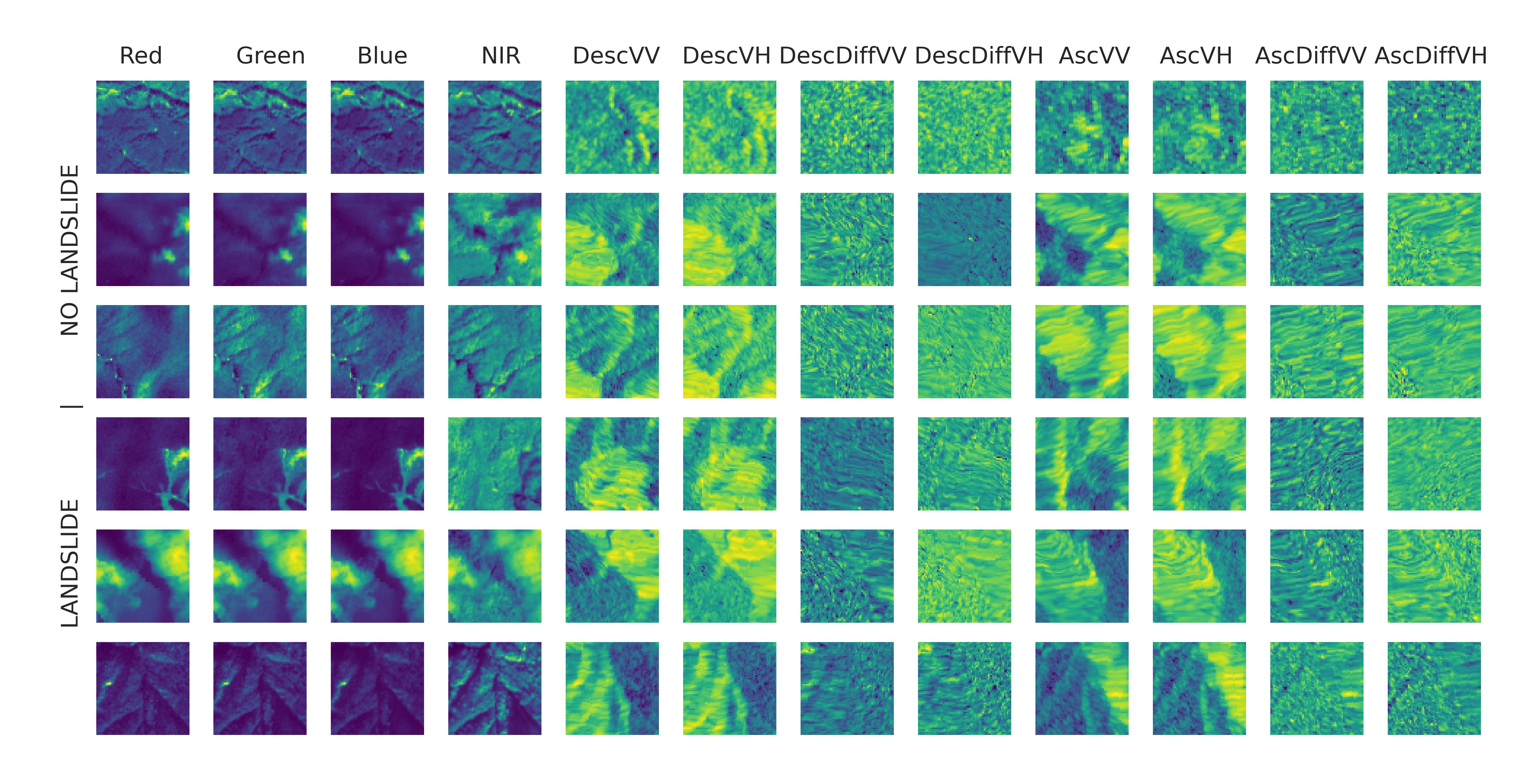}
    \caption{All bands plots for 6 samples}
    \label{fig:samplesNbands}
\end{figure}

\begin{figure}[h]
    \centering
    \includegraphics[width=0.55\textwidth]{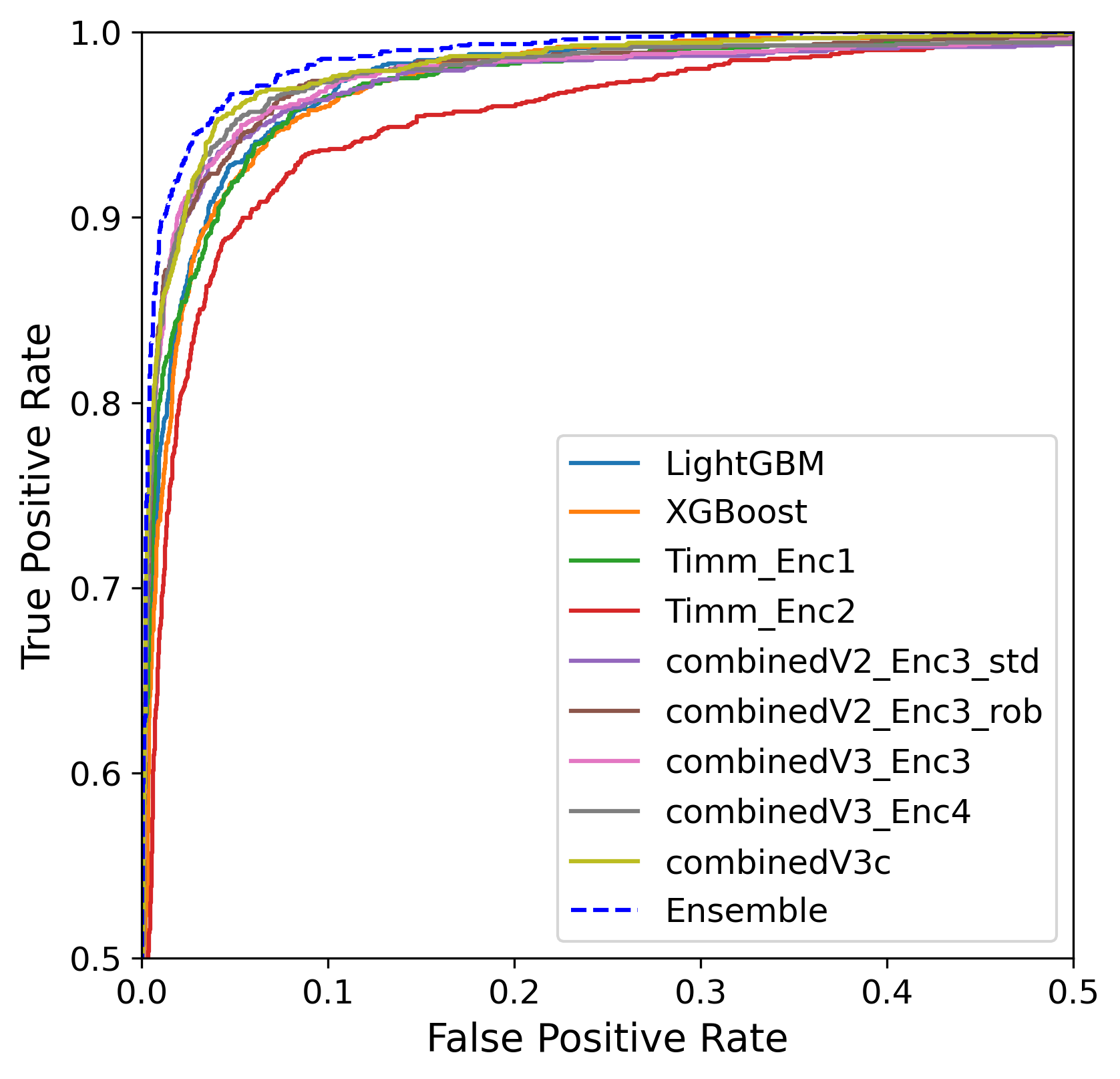}
    \caption{ROC curves}
    \label{fig:ROCcurve}
\end{figure}

\autoref{fig:samplesNbands} presents six representative 64 $\times$ 64 patches across all bands and \autoref{fig:ROCcurve} depicts the ROC curves for every model used. 

\newpage
\section{Other Experiments}
\label{sec:Appendix B}

Initial experiments with timm-based models using the high-performing RGBN modality evaluated several well-established architectures, widely adopted both in EO and in general computer vision tasks (\autoref{table:model_scoresRGBN}). These trials highlighted the superior performance of transformer-based architectures, with MaxVit achieving the highest scores (\autoref{table:model_scores}). Subsequent investigations into scaling strategies (\autoref{table:model_scoresScaling}) demonstrated that robust scaling using the 1st-99th percentiles produced results worse than those obtained with the 5th-95th percentile range, leading to their exclusion in later trials. Toward the end of the experimentation phase, more advanced evaluations confirmed the consistent superiority of transformers and emphasized the benefit of employing the same encoder type across different modalities (\autoref{table:model_scoresV}). Finally, replacing the lighter Enc3 encoder (\autoref{table:model_scores}) with the heavier Enc1 did not yield any performance gains (\autoref{table:model_scoresV}), highlighting the effectiveness and efficiency of the lightweight encoder design.

\begin{table}[h]
\caption{Exploratory experiments without vision transformers}
\label{table:model_scoresRGBN}
\centering
    \addtolength{\leftskip} {-4cm}
    \addtolength{\rightskip}{-4cm}
\begin{tabular}{l  l  cccc l l l l} 
 \hline
 Model & Encoder & RGBN & SAR & SAR diff & Indices & Scaling & OOF & pu/pr LB & Overall\\ 
 \hline
Timm &  convnextv2\_tiny.fcmae\_ft\_in22k\_in1k &  \checkmark & & &  & Robust & 0.8716 & 0.862/0.8541 & 0.864 \\ 
Timm &  edgenext\_base.in21k\_ft\_in1k	 &  \checkmark & & &  & Robust & 0.8748 & 0.8786/0.8554 & 0.8686 \\ 
Timm &  tf\_efficientnet\_b3\_ns &  \checkmark & & &  & Robust & 0.8292 & 0.8463/0.823 & 0.8296 \\ 


\end{tabular}

\end{table}

\begin{table}[h]
\caption{Exploratory experiments with different scalings}
\label{table:model_scoresScaling}
\centering
    \addtolength{\leftskip} {-4cm}
    \addtolength{\rightskip}{-4cm}
\begin{tabular}{l  l  cccc l l l l} 
 \hline
 Model & Encoder & RGBN & SAR & SAR diff & Indices & Scaling & OOF & pu/pr LB & Overall\\ 
 \hline
combinedV2 &  edgenext\_small.usi\_in1k &  \checkmark & & \checkmark &  & Standard & 0.8763 & 0.8704/0.8565 & 0.8685 \\ 
combinedV2 &  edgenext\_small.usi\_in1k &  \checkmark & & \checkmark &  & Robust (1, 99) & 0.8753 & 0.839/0.854 & 0.8624 \\ 
combinedV2 &  edgenext\_small.usi\_in1k &  \checkmark & & \checkmark &  & Robust (5, 95)& 0.8781 & 0.8894/0.8656 & 0.8754 \\ 


\end{tabular}

\end{table}

\begin{table}[h]
\caption{Other experiments}
\label{table:model_scoresV}
\centering
    \addtolength{\leftskip} {-4cm}
    \addtolength{\rightskip}{-4cm}
\begin{tabular}{l  l  cccc l l l l} 
 \hline
 Model & Encoder & RGBN & SAR & SAR diff & Indices & Scaling & OOF & pu/pr LB & Overall\\ 
 \hline
combinedV2 &  Enc1 &  \checkmark & & \checkmark &  & Standard & 0.8953 & 0.9169/0.8854 & 0.8951 \\ 
combinedV3c &  edgenext\_small.usi\_in1k &  \checkmark & & \checkmark &  \checkmark & Standard & 0.8862 & 0.8606/0.8668 & 0.8756 \\ 
combinedV2 &  Enc1 and Enc3 &  Enc1 & & Enc3 &  & Robust & 0.8898 & 0.9037/0.8818 & 0.8891 \\ 
 

\end{tabular}

\end{table}

\end{document}